\documentclass[cameraready]{Interspeech}

\usepackage{fontspec}
\usepackage{xeCJK}
\usepackage{tipa}
\usepackage{xcolor}

\newfontfamily\ipafont{FreeSerif.otf}

\setCJKsansfont{NotoSerifCJKsc-Regular.otf}[
  Path=fonts/,
  BoldFont=NotoSerifCJKsc-Regular.otf
]

\setCJKmonofont{NotoSerifCJKsc-Regular.otf}[
  Path=fonts/
]

\setCJKfamilyfont{korean}[
  Path=fonts/
]{NotoSansCJKkr-Regular.otf}

\newcommand{\ko}[1]{{\CJKfamily{korean}#1}}

\newfontfamily\thaifont[
  Path=fonts/,
  Extension=.ttf,
  UprightFont=*-Regular,
  BoldFont=*-Bold,
  Script=Thai
]{NotoSerifThai}

\newfontfamily\laofont[
  Path=fonts/,
  Extension=.ttf,
  UprightFont=*-Regular,
  BoldFont=*-Bold,
  Script=Lao
]{NotoSerifLao}

\newcommand{\thai}[1]{{\thaifont #1}}
\newcommand{\lao}[1]{{\laofont #1}}

\title{A Phonemically Comprehensive, ASCII-Only Romanization Scheme for Thai and Lao: Systematic Cross-Lingual Correspondence and Chinese-User-Friendly Design}

\author[affiliation={1}]{Zijie}{Zhang}
\author[affiliation={1}, correspondingauthor]{Tan}{Lee}

\address{
    $^1$ The Chinese University of Hong Kong, Shenzhen, China
}

\email{zijiezhang@link.cuhk.edu.cn, tanlee@cuhk.edu.cn}


\usepackage{comment}

\begin{document}

\maketitle

\begin{abstract}
    This paper proposes a phonemically comprehensive, ASCII-only romanization scheme for Thai and Lao, treating the two closely related languages as a unified cross-lingual design problem. The scheme represents segmental contrasts, vowel length, and lexical tone while maintaining one-symbol-one-phoneme transparency and systematic correspondence between Thai and Lao. The scheme prioritizes synchronic phonetic correspondence, including correspondence with Pinyin and Jyutping where applicable, while preserving historical-phonological correspondence where it does not conflict with phonetic transparency. Tone uses a compact single-digit default notation, supplemented by optional tone-value and historical tone-category representations. The resulting scheme provides a readable, keyboard-friendly, and machine-processable phonemic representation for language learning and cross-lingual speech processing.
\end{abstract}

\section{Introduction}

Romanization, broadly defined, refers to the representation of a language
originally written in a non-Latin script using the Latin alphabet.
In the context of this paper, the term is used more specifically to denote
a systematic, phonologically faithful Latin-script encoding of the
\textit{pronunciation} of Thai and Lao---two closely related languages
of the Southwestern branch of the Tai--Kadai (Kra--Dai) family.
Such a representation is not merely a transliteration of graphemes,
but a transcription that aims to preserve the phonological contrasts relevant to pronunciation of the target language, including segmental contrasts, vowel length,
and lexical tone.

A major motivation for a Thai–Lao romanization scheme is the difficulty that their complex writing systems pose for non-native learners. Empirical studies of Thai as a foreign language have repeatedly identified the writing system itself as a primary source of difficulty.
Hou \cite{hou2017spelling} found that the complexity of the Thai writing system was a major cause of spelling errors among Chinese learners, outweighing simple L1 phonological interference in many cases.
Elliott \cite{elliott2012l2ws} similarly argues that features such as non-linear vowel placement, consonant-class tone encoding, and the absence of regular word spacing in Lao (and by extension Thai) create particular decoding difficulties for adult learners whose L1 writing systems lack these properties.
Classroom observations reported that many L2 learners of Thai ultimately abandon the script altogether and confine their learning to the spoken language, in contrast to the relative ease with which learners acquire more transparent systems such as Hangul or kana \cite{y14-1038}.
Practical language-learning and travel resources likewise consistently describe the Thai and Lao scripts as a genuine barrier to rapid acquisition of pronunciation for short-term visitors and independent learners \cite{jamkham2026,pasaa2026,lingapp2026,expatden2026}.

Beyond facilitating human learning, a well-designed romanization scheme for Thai and Lao is valuable for computational applications.
In particular, grapheme-to-phoneme (G2P) conversion has long been recognized as a crucial front-end component of Thai text-to-speech (TTS) systems \cite{rugchatjaroen2019efficient,geng2025bridging}.
Because Thai orthography lacks word boundaries, encodes tone through a complex interaction of consonant class, syllable type and diacritics, and retains numerous etymological spellings, accurate conversion from written form to a phonemic representation is indispensable for high-quality synthesis.
Such a romanization scheme can therefore serve directly as the phonemic tier (the ``P'' in G2P), providing an intermediate representation that is both machine-readable and human-interpretable. Lao, as a low-resource language with severely limited publicly available speech data \cite{liu2025lao,geng2025bridging}, may particularly benefit from such a representation.
In principle, a romanization scheme that maintains consistent cross-lingual phonetic correspondence could facilitate transfer learning from the relatively better-resourced Thai to Lao, thereby alleviating data scarcity \cite{shurtz2025scripts,wells2021cross}. The combination of human readability and computational utility underscores the practical value of a well-designed Thai--Lao romanization scheme.

The present paper proposes a new romanization scheme for Thai and Lao.
Rather than treating the two languages as independent design problems, we approach the task as a single cross-lingual engineering effort.
This unified scheme considers both systematic Thai–Lao correspondence and phonetic correspondence with established external romanization schemes, most notably Pinyin and Jyutping.
The scheme is designed according to the following principles:
\begin{enumerate}
\item \textbf{Comprehensive phonemic representation}
\item \textbf{Strict ASCII-only inventory}
\item \textbf{Chinese-user-friendly design via phonetic correspondence with Pinyin and Jyutping}
\item \textbf{Alphanumeric design for readability}
\item \textbf{One-symbol-one-phoneme transparency}
\item \textbf{Cross-lingual phonetic correspondence between Thai and Lao}
\item \textbf{Historical-phonological correspondence}
\item \textbf{Priority of synchronic phonetic correspondence}
\end{enumerate}

The proposed romanization scheme maps directly to the phonemic systems of Thai and Lao rather than to their traditional orthographies. The original orthographic forms therefore cannot be uniquely recovered from the romanized form.

The remainder of the paper is organized as follows.
Section~\ref{sec:principles} elaborates and justifies each of the eight principles in detail.
Section~\ref{sec:existing-schemes} reviews existing romanization schemes for Thai and Lao and identifies the limitations that motivate the present scheme.
Section~\ref{sec:proposed-scheme} then presents the proposed scheme, explaining and motivating its concrete symbol choices in light of the principles articulated above.

\section{Design Principles}
\label{sec:principles}

\subsection{Comprehensive Phonemic Representation}

The first principle requires that the scheme encode the phonological contrasts adopted in the present analyses of Thai and Lao, including segmental contrasts, vowel length, and lexical tone.

A phonemically incomplete representation fails on two fronts simultaneously.
For human learners, any omitted contrast (for example, the failure to distinguish long versus short vowels, or the collapse of aspirated and unaspirated stops) removes the very information needed to recover correct pronunciation from the written form; the learner is forced to rely on external knowledge or guesswork \cite{wells1995sampa}.
For speech technology, the same omission is equally damaging.
Modern Thai TTS pipelines treat an accurate phonemic string—complete with tone and length—as the essential intermediate representation that drives the acoustic model \cite{rugchatjaroen2019efficient,geng2025bridging}.
When a romanization scheme systematically under-specifies the phonology, it can no longer serve as a reliable “P” in G2P conversion, nor can it support cross-lingual transfer that depends on shared phonemic inventories \cite{shurtz2025scripts}.

Thus comprehensive phonemic coverage is not an optional refinement but the minimal condition for a romanization scheme that is usable both by language learners and by computational systems.

\subsection{Strict ASCII-only Inventory}

The second principle requires all symbols used by the scheme to be drawn from ASCII.

This restriction is motivated by long-standing practical considerations in both human and machine use of phonetic transcription.
Systems such as SAMPA and X-SAMPA were deliberately designed as 7-bit ASCII encodings of the IPA precisely because non-ASCII phonetic symbols create persistent obstacles: they are difficult to type on ordinary keyboards, frequently lack font support, and introduce encoding errors in plain-text environments, electronic mail, and legacy databases \cite{wells1995sampa,kirshenbaum1992}.
An ASCII-only representation avoids these barriers.
It can be entered on standard keyboards without specialized input methods and processed as ordinary text for sorting, searching, tokenization, and dictionary lookup.

\subsection{Phonetic Correspondence with Pinyin and Jyutping}

China maintains extensive people-to-people exchanges with Thailand and Laos.
Official statistics show that Chinese travelers constitute one of the largest groups of international arrivals in both countries, while cross-border movement between the Chinese mainland and ASEAN states exceeded 25 million person-times in the first eight months of 2025 alone \cite{nia2025asean,laotourism2024}.
A romanization scheme that is immediately accessible to users familiar with Pinyin or Jyutping therefore has the potential to benefit a substantial population of learners, travelers and professionals.

Shared script–sound knowledge between a familiar system and a target system facilitates the reuse of prior knowledge and accelerates acquisition of the new system \cite{odlin1989}.
Pinyin is the official and most widely used romanization scheme for Standard Chinese in China, and has been adopted by the United Nations and the International Organization for Standardization \cite{pinyin1958,iso7098}.
Jyutping, promulgated by the Linguistic Society of Hong Kong in 1993, is one of the most authoritative and systematically designed romanization schemes currently employed for Cantonese—one of the most influential Sinitic languages in southern China \cite{lshk1993,jyutping-scheme}.

Consequently, we regard systematic \textit{phonetic correspondence} with both Pinyin and Jyutping as a core design principle:
phonemes that possess similar or identical phonetic realizations across the languages in question should be represented by similar or identical romanized symbols~\cite{zhang2026iscslp}.
Aligning the Thai–Lao scheme with these two established Chinese romanization schemes facilitates positive transfer for users familiar with Pinyin or Jyutping while preserving phonological transparency. 

\subsection{Alphanumeric design for readability}

Although an ASCII-only inventory already improves input convenience and human–computer interaction, not every ASCII character is equally suitable for a readable romanization scheme.
Certain punctuation symbols that fall inside the ASCII range—most notably \texttt{?}, \texttt{\^{}} and \texttt{\~{}}—have been employed in legacy transcription systems such as Vietnamese Quoted-Readable (VIQR) to stand in for diacritics \cite{viqr}.
These symbols, however, do not function as ordinary alphabetic symbols in familiar romanized orthographies; their presence therefore produces visually alien strings that impede rapid reading and intuitive recognition.

Neither Pinyin nor Jyutping relies on ASCII punctuation marks such as \texttt{?}, \texttt{\^{}} or \texttt{\~{}} as core phonological symbols.
We therefore restrict the core inventory of the present scheme to the same characters: \texttt{a-z0-9}.
Upper-case letters \texttt{A-Z} appear only in an optional multi-track tone notation that encodes historical class or etymological source; this device is an ancillary convenience rather than part of the primary orthographic convention and will be described in detail later.

\subsection{One-Symbol-One-Phoneme Transparency}

The fifth principle requires that each phoneme be represented by a unique and unambiguous symbol. Here, a symbol may consist of a single letter, a digraph, or a longer multigraph functioning as a single graphemic unit. In the design of Jyutping this is stated explicitly as the requirement that “One symbol to one phoneme mapping” (\textit{一個符號對應一個音位}), so that the sound of a symbol can be determined without reference to its orthographic environment \cite{jyutping-scheme}.
Within the core phonemic representation, allowing the same symbol to represent different phonemes according to context, or assigning alternative symbols to the same phoneme, would introduce unnecessary ambiguity for both human learners and computational processing.
By maintaining a strict one-to-one mapping we obtain a representation that is clear, accurate and concise, and that remains highly predictable from the written form alone.

\subsection{Cross-lingual Phonetic Correspondence between Thai and Lao}

Thai and Lao belong to the Southwestern branch of the Tai–Kadai family and exhibit extensive phonological overlap.
In particular, their vowel systems are nearly identical in both inventory and phonetic realization (the classic nine-vowel system with length contrast), and their coda inventories are almost the same.
Consequently, a large proportion of the phonemes in the two languages have essentially identical or extremely close phonetic values.

The sixth principle therefore requires that the scheme assign the same (or highly similar) symbols to these shared or near-identical phonemes with identical or near-identical phonetic realizations across the two languages.
This systematic cross-lingual correspondence yields two practical benefits.
For human learners, mastery of the symbol-to-sound mappings in one language supports reliable prediction of the other.
For computational applications, the shared representation can theoretically facilitate parameter sharing and cross-lingual transfer from the relatively higher-resource Thai to low-resource Lao speech processing tasks \cite{shurtz2025scripts,liu2025lao}.

\subsection{Historical-Phonological Correspondence}

The seventh principle concerns diachronic relatedness.
For historically corresponding phonemes, the scheme prefers identical or highly similar symbols~\cite{zhang2026iscslp}.
As a result, cognate words across Thai and Lao tend to receive identical or near-identical spellings.

This design choice yields benefits for both human learners and computational systems.
On the learning side, the spelling similarity of cognates allows a learner who has mastered the romanization of one language to transfer knowledge more readily to the other—an advantage consistent with the well-established finding that cognates are acquired and retrieved more easily than non-cognates \cite{lotto1998effects}.
Computationally, mapping historically related phonemes to similar symbols makes cognates more detectable by string-similarity metrics and supports cross-lingual transfer, out-of-vocabulary handling, and parameter sharing, in the same spirit as universal romanization tools that deliberately preserve cognate similarity across scripts \cite{hermjakob2018uroman}.

\subsection{Priority of Synchronic Phonetic Correspondence}

Historically corresponding forms may diverge substantially in their modern pronunciations, while phonetically similar forms may have different historical origins. The present scheme is not an attempt to reconstruct a common ancestral form.
Its primary purpose is synchronic readability and learnability: within each language the symbol-to-sound mappings must remain transparent and intuitive to learners.

Only secondarily does the scheme seek to reflect historical relatedness.
Consequently, when phonetic correspondence (Principle~6) and historical-phonological correspondence (Principle~7) come into conflict, the former is given precedence.
We do not force identical or similar spellings for cognates at the cost of rendering the symbol-to-sound relationship opaque or counter-intuitive in either language.

\section{Existing Thai and Lao Romanization Schemes}
\label{sec:existing-schemes}

A number of romanization schemes have been proposed or adopted for Thai and Lao. We briefly evaluate the major ones against our design principles.

For Thai, the official Royal Thai General System of Transcription
(RTGS) is strictly ASCII and highly readable, yet deliberately omits
tone and vowel length and collapses several segmental contrasts; it
therefore does not provide comprehensive phonemic representation
~\cite{kanchanawan2006romanization}.
ISO 11940 and the ALA-LC system achieve near-complete orthographic
reversibility but rely on extensive diacritics, rendering them neither
ASCII-only nor particularly readable for non-specialists
~\cite{ISO11940,loc2011thai}.
ISO 11940, in particular, is fundamentally an orthographic
transliteration rather than a direct phonemic transcription: it
transliterates Thai characters individually, preserves distinctions
among orthographic symbols that may share the same phonemic
realization, and retains orthographic material that is not itself
pronounced~\cite{ISO11940}. Although tone marks and the distinctions
among consonant classes are recoverable from the transliteration, the
lexical tone is not represented directly; the reader must still derive
it from the interaction of consonant class, tone marking, syllable
type, and vowel length~\cite{iwasaki2005reference}. Consequently,
recovering pronunciation from ISO 11940 still presupposes substantial
knowledge of Thai orthographic and phonological rules rather than
being immediately transparent from the romanized form alone.
Learner-oriented systems such as Paiboon (and its later variants) mark tones and length, yet employ
non-ASCII symbols (e.g., {\ipafont ʉ}, tone diacritics)
~\cite{becker2009threeway}. Among widely used Thai schemes, the Enhanced Phonemic notation of thai-language.com (TLC) is notable for satisfying the constraints: it remains within ASCII and marks all phonemic contrasts, including tone through trailing M/L/H/F/R tags and vowel length through doubling.

For Lao the situation is similar. The BGN/PCGN 1966 system uses a small number of non-ASCII diacritics and does not represent lexical tone or vowel length, while the more recent Ministry of Health 2020 system is ASCII-only but likewise does not distinguish tone or vowel length. ALA-LC again introduces diacritics~\cite{loc2012lao}. The input method known as LaoScript comes closest to comprehensive phonemic coverage and is essentially ASCII-based; however, it is fundamentally an
orthography-oriented input transliteration rather than a direct
phonemic transcription~\cite{durdin2021laoscript}. It preserves
distinctions among orthographic symbols that may correspond to the
same segmental phoneme, accommodates orthographic material that is
not itself pronounced, and represents tone marks rather than lexical
tone directly~\cite{durdin2021laoscript,lew2014lao}. Consequently,
the actual lexical tone must still be derived from the interaction
of consonant class, tone marking, and other syllable properties, so
recovering pronunciation requires knowledge of Lao orthographic
rules rather than being immediately transparent from the romanized
form alone.

Crucially, none of the systems reviewed above was designed as a
unified romanization scheme for both Thai and Lao, nor do they incorporate
systematic cross-lingual correspondence between the two languages.
This absence of cross-lingual design leaves a clear gap that the
present scheme aims to fill.

\section{The Proposed Romanization Scheme}
\label{sec:proposed-scheme}

\subsection{Initial Consonant System}
\label{sec:initials}

We begin by presenting the initial consonant inventories of Thai and Lao together with their corresponding romanization symbols.
Shared phonemes that have identical or near-identical phonetic realizations in both languages are assigned the same symbol, in accordance with the principle of cross-lingual phonetic correspondence.
Phonemes unique to one language are listed only in the relevant column.

\begin{table}[ht]
\centering
\caption{Initial consonants and their romanization}
\label{tab:initials}
\begin{tabular}{@{}p{5.2cm}cc@{}}
\toprule
\textbf{Phoneme (IPA)} & \textbf{Thai} & \textbf{Lao} \\
\midrule
{\ipafont /p/}                          & \multicolumn{2}{c}{$\langle$b$\rangle$} \\
{\ipafont /p\textsuperscript{h}/}       & \multicolumn{2}{c}{$\langle$p$\rangle$} \\
{\ipafont /b/}                          & \multicolumn{2}{c}{$\langle$bb$\rangle$} \\
{\ipafont /t/}                          & \multicolumn{2}{c}{$\langle$d$\rangle$} \\
{\ipafont /t\textsuperscript{h}/}       & \multicolumn{2}{c}{$\langle$t$\rangle$} \\
{\ipafont /d/}                          & \multicolumn{2}{c}{$\langle$dd$\rangle$} \\
{\ipafont /k/}                          & \multicolumn{2}{c}{$\langle$g$\rangle$} \\
{\ipafont /k\textsuperscript{h}/}       & \multicolumn{2}{c}{$\langle$k$\rangle$} \\
{\ipafont /tɕ/}                         & \multicolumn{2}{c}{$\langle$z$\rangle$} \\
{\ipafont /tɕ\textsuperscript{h}/}      & $\langle$ch$\rangle$ & --- \\
{\ipafont /m/}                          & \multicolumn{2}{c}{$\langle$m$\rangle$} \\
{\ipafont /n/}                          & \multicolumn{2}{c}{$\langle$n$\rangle$} \\
{\ipafont /ŋ/}                          & \multicolumn{2}{c}{$\langle$ng$\rangle$} \\
{\ipafont /ɲ/}                          & ---                  & $\langle$nj$\rangle$ \\
{\ipafont /f/}                          & \multicolumn{2}{c}{$\langle$f$\rangle$} \\
{\ipafont /s/}                          & \multicolumn{2}{c}{$\langle$s$\rangle$} \\
{\ipafont /h/}                          & \multicolumn{2}{c}{$\langle$h$\rangle$} \\
{\ipafont /r/}                          & $\langle$r$\rangle$  & --- \\
{\ipafont /l/}                          & \multicolumn{2}{c}{$\langle$l$\rangle$} \\
{\ipafont /j/}                          & \multicolumn{2}{c}{$\langle$j$\rangle$} \\
{\ipafont /w/} {\footnotesize (initial; also Cw cluster or labialized)} 
                                        & \multicolumn{2}{c}{$\langle$w$\rangle$} \\
{\footnotesize zero onset, optionally realized as} {\ipafont [ʔ]}                    & \multicolumn{2}{c}{$\langle$gh$\rangle$ (optional)} \\
\bottomrule
\end{tabular}
\end{table}

The onset consonant inventory of the proposed scheme is designed
to satisfy the eight design principles introduced before. The design decisions are discussed below.

\noindent\textbf{Pinyin- and Jyutping-inspired voiceless stop onsets.}
Thai and Lao both exhibit a three-way contrast in the stop series:
voiceless unaspirated, voiceless aspirated, and voiced---the last being
phonetically realized as implosives {\ipafont [ɓ]} and {\ipafont [ɗ]} in both
languages~\cite{iwasaki2005reference, enfield2007grammar, noss1964thai}.
The proposed scheme represents the voiceless unaspirated stops
{\ipafont /p, t, k/} as $\langle\mathrm{b, d, g}\rangle$, their aspirated counterparts
{\ipafont /pʰ, tʰ, kʰ/} as $\langle\mathrm{p, t, k}\rangle$, and the voiced stops
{\ipafont /b, d/}, phonetically realized as {\ipafont [ɓ, ɗ]}, as
$\langle\mathrm{bb, dd}\rangle$.

The mapping of voiceless unaspirated stops to $\langle\mathrm{b, d, g}\rangle$
and voiceless aspirated stops to $\langle\mathrm{p, t, k}\rangle$ follows the
established conventions of both Pinyin~\cite{norman1988chinese} and
Jyutping~\cite{bauer1997cantonese, matthews2011cantonese},
thereby reducing the learning burden for speakers of Mandarin
or Cantonese. The implosive series {\ipafont /ɓ, ɗ/} is represented by
doubled graphemes $\langle\mathrm{bb, dd}\rangle$, a visually transparent
extension of the basic $\langle\mathrm{b, d}\rangle$ symbols that signals
a phonologically marked category without introducing any non-ASCII
character.

\noindent\textbf{\boldmath$\langle\mathrm{j}\rangle$ for {\ipafont /j/}.}
The voiced palatal approximant {\ipafont /j/} is represented by
$\langle\mathrm{j}\rangle$, consistent with the Jyutping convention~\cite{matthews2011cantonese}.
The alternative grapheme $\langle\mathrm{y}\rangle$ is deliberately avoided.
In both Pinyin and Jyutping, the sequence
$\langle\mathrm{yu}\rangle$ is associated with the close front rounded vowel
{\ipafont /y/}
(e.g., Pinyin \textit{yu2} {\ipafont /y³⁵/} `fish';
Jyutping \textit{jyu4} {\ipafont /y²¹/} `fish')
~\cite{norman1988chinese, matthews2011cantonese}.
If {\ipafont /j/} were encoded as $\langle\mathrm{y}\rangle$,
a Thai or Lao syllable such as {\ipafont /juː/}
(Thai \thai{อยู่}, Lao \lao{ຢູ່} `to be at') would be romanized as
$\langle\mathrm{yu}\rangle$, which could readily be misread by
Pinyin- and Jyutping-literate users as {\ipafont /y/} rather than
{\ipafont /juː/}.
This potential source of negative transfer is avoided by reserving
$\langle\mathrm{j}\rangle$ exclusively for the palatal approximant.

\noindent\textbf{\boldmath$\langle\mathrm{z}\rangle$ for {\ipafont /tɕ/} and
$\langle\mathrm{ch}\rangle$ for {\ipafont /tɕʰ/}.}
The voiceless alveolo-palatal affricate {\ipafont /tɕ/} is represented by
$\langle\mathrm{z}\rangle$. 
From the perspective of Pinyin, the more directly corresponding choice $\langle\mathrm{j}\rangle$ is unavailable, as it is already assigned to {\ipafont /j/}. The selection of $\langle\mathrm{z}\rangle$ is motivated by the following phonetic correspondences:

\begin{enumerate}
\item In Jyutping, $\langle\mathrm{z}\rangle$ represents {\ipafont /ts/}, with the postalveolar realization {\ipafont [tʃ]} also occurring in Cantonese~\cite{matthews2011cantonese,bauer1997cantonese}. The postalveolar {\ipafont [tʃ]} and alveolo-palatal {\ipafont [tɕ]} are closely related articulations: alveolo-palatal consonants have been characterized as palatalized postalveolars, differing from palato-alveolar articulations principally in the location of constriction and a greater degree of tongue-dorsum raising and palatalization~\cite{ladefoged1996sounds,recasens2013alveolopalatal}.

\item In Pinyin, $\langle\mathrm{z}\rangle$ represents the voiceless unaspirated alveolar affricate {\ipafont /ts/}. Both {\ipafont /ts/} and {\ipafont /tɕ/} are voiceless unaspirated coronal sibilant affricates, differing primarily in the location of the coronal constriction and the degree of palatalization~\cite{ladefoged1996sounds}. Since Thai and Lao do not phonemically contrast {\ipafont /ts/} with {\ipafont /tɕ/}~\cite{iwasaki2005reference,enfield2007grammar}, assigning $\langle\mathrm{z}\rangle$ to {\ipafont /tɕ/} introduces no within-system ambiguity.
\end{enumerate}

For the voiceless aspirated alveolo-palatal affricate
{\ipafont /tɕʰ/}, the proposed scheme uses
$\langle\mathrm{ch}\rangle$ rather than the Pinyin
$\langle\mathrm{q}\rangle$. The rationale is threefold:

\begin{enumerate}
\item The grapheme $\langle\mathrm{q}\rangle$ is highly non-intuitive
for users unfamiliar with Pinyin. For readers accustomed to English
and similar Latin-script orthographic conventions,
$\langle\mathrm{q}\rangle$ is not ordinarily associated with an affricate,
making the Pinyin mapping $\langle\mathrm{q}\rangle$--{\ipafont /tɕʰ/}
particularly counterintuitive~\cite{wheatley2011learning,hayesharb2016pinyin}.

\item In Pinyin, $\langle\mathrm{ch}\rangle$ represents the voiceless
aspirated retroflex affricate {\ipafont /tʂʰ/}. The retroflex
{\ipafont /tʂʰ/} and alveolo-palatal {\ipafont /tɕʰ/} are phonetically
and perceptually similar voiceless aspirated coronal sibilant affricates,
sharing the same manner and laryngeal specification and differing
principally in place of articulation
\cite{yu2023mandarinsimilarity,rasmussen2015mandarin,chen2023multilanguage}.
This distinction is not phonemically contrastive in Thai or Lao
\cite{iwasaki2005reference,enfield2007grammar}.

\item Although Jyutping itself does not use the digraph
$\langle\mathrm{ch}\rangle$, it is familiar in conventional Hong Kong
Cantonese romanization schemes, where it represents {\ipafont /tsʰ/}, which
may have a postalveolar realization {\ipafont [tʃʰ]}
\cite{matthews2011cantonese,bauer1997cantonese}.
The postalveolar {\ipafont [tʃʰ]} and alveolo-palatal
{\ipafont /tɕʰ/} are articulatorily close, differing primarily in
the location of the coronal constriction and degree of palatalization
\cite{ladefoged1996sounds,recasens2013alveolopalatal}.
Thus $\langle\mathrm{ch}\rangle$ provides a familiar and phonetically
motivated representation of an aspirated affricate for Cantonese-speaking users.
\end{enumerate}

\noindent\textbf{\boldmath$\langle\mathrm{nj}\rangle$ for {\ipafont /ɲ/}.}
The voiced palatal nasal {\ipafont /ɲ/} is present in Lao but absent from
Thai~\cite{enfield2007grammar}. In the proposed scheme, it is
represented by $\langle\mathrm{nj}\rangle$. This choice is motivated by
both synchronic phonetics and diachronic correspondence:

\begin{enumerate}
\item \textbf{Synchronic motivation.}
The palatal nasal {\ipafont /ɲ/} and the palatal approximant {\ipafont /j/}
share the same place of articulation (palatal). The symbol
$\langle\mathrm{nj}\rangle$ can be parsed compositionally as
$\langle\mathrm{n}\rangle$ (indicating nasality) +
$\langle\mathrm{j}\rangle$ (indicating palatal place), yielding a
transparent representation of the palatal nasal
~\cite{ladefoged1996sounds}.

\item \textbf{Diachronic motivation.}
In Proto-Tai, the onset {\ipafont */ɲ/} is reconstructed for a set of
cognate lexical items. In the Southwestern Tai branch, this
phoneme has been preserved as {\ipafont /ɲ/} in Lao
(e.g., \lao{ຍຸງ} {\ipafont /ɲuŋ/} `mosquito') but has undergone
denasalization to {\ipafont /j/} in Thai
(e.g., \thai{ยุง} {\ipafont /juŋ/} `mosquito')
~\cite{li1977handbook}.
By representing Lao {\ipafont /ɲ/} as $\langle\mathrm{nj}\rangle$
and Thai {\ipafont /j/} as $\langle\mathrm{j}\rangle$, the scheme
preserves a visual trace of this historical correspondence:
$\langle\mathrm{nj}\rangle$ is transparently related to
$\langle\mathrm{j}\rangle$ by the addition of the nasal prefix
$\langle\mathrm{n}\rangle$.
This exemplifies the design principle of
\textit{historical-phonological correspondence}
discussed in Section~\ref{sec:principles}.
\end{enumerate}

\noindent\textbf{Onset clusters and labialization.}
Onset clusters are romanized compositionally: each consonantal
element retains the romanization assigned to the corresponding
singleton onset, and the resulting graphemes are concatenated in the
same phonological order. Thus, for example, Thai {\ipafont /pr/} is
written $\langle\mathrm{br}\rangle$, combining
$\langle\mathrm{b}\rangle$ for {\ipafont /p/} with
$\langle\mathrm{r}\rangle$ for {\ipafont /r/}; the same principle
applies systematically to the other Thai onset clusters
~\cite{iwasaki2005reference}.

The labial component in Lao velar onsets is analysis-dependent. Sequences conventionally represented as
{\ipafont /kw/} and {\ipafont /kʰw/} may alternatively be analyzed
as the labialized consonants {\ipafont /kʷ/} and
{\ipafont /kʰʷ/}~\cite{erickson2001labialized,enfield2007grammar}.
The proposed scheme is deliberately neutral with respect to
this distinction: $\langle\mathrm{w}\rangle$ represents either the
second {\ipafont /w/} element of a cluster or the secondary
labialization feature {\ipafont /ʷ/}. Consequently, both analyses
yield the same spellings, $\langle\mathrm{gw}\rangle$ and
$\langle\mathrm{kw}\rangle$, respectively. Thai
{\ipafont /kw, kʰw/}, which are ordinarily analyzed as onset
clusters, receive the same representations by the general
compositional rule~\cite{iwasaki2005reference}. The resulting
$\langle\mathrm{gw}\rangle$--$\langle\mathrm{kw}\rangle$ pair is
therefore shared across Thai and Lao independently of the preferred
phonological analysis, while also coinciding with the established
Jyutping spellings for the Cantonese labialized velar onsets
{\ipafont /kʷ, kʷʰ/}~\cite{bauer1997cantonese,matthews2011cantonese}.

\noindent\textbf{\boldmath Optional $\langle\mathrm{gh}\rangle$ for zero onset.}
Thai and Lao do not exhibit a phonemic contrast between the glottal
stop {\ipafont /ʔ/} and a true zero onset ($\varnothing$). In both languages, syllables with a phonological zero onset can be phonetically realized with an initial glottal stop, but this glottal
stop is an automatic phonetic insertion rather than a distinctive
phoneme~\cite{iwasaki2005reference, enfield2007grammar}.

As an optional phonetic device outside the core phonemic mapping,
$\langle\mathrm{gh}\rangle$ may be used to represent this inserted glottal stop.
The design rationale is phonetically motivated:

\begin{enumerate}
\item The grapheme $\langle\mathrm{g}\rangle$ is already used to represent
the voiceless velar plosive {\ipafont /k/}. The glottal stop
{\ipafont /ʔ/} shares the same \textit{manner} of articulation
(complete closure, i.e., stop/plosive) as {\ipafont /k/},
differing only in \textit{place} (glottal versus
velar)~\cite{ladefoged1996sounds}.

\item The grapheme $\langle\mathrm{h}\rangle$ represents the voiceless
glottal fricative {\ipafont /h/}. The glottal stop {\ipafont /ʔ/}
shares the same \textit{place} of articulation (glottal) as
{\ipafont /h/}, differing only in \textit{manner} (stop versus
fricative)~\cite{ladefoged1996sounds}.

\item By combining $\langle\mathrm{g}\rangle$ (stop manner) and
$\langle\mathrm{h}\rangle$ (glottal place), the symbol
$\langle\mathrm{gh}\rangle$ provides a phonetically motivated,
purely ASCII representation of the glottal stop.
\end{enumerate}

However, $\langle\mathrm{gh}\rangle$ is entirely optional. In the default
representation, syllables with a phonological zero onset but a phonetic glottal stop onset can be written with the vowel symbol directly (e.g.,
$\langle\mathrm{aa}\rangle$ for {\ipafont [ʔaː]}).
The $\langle\mathrm{gh}\rangle$ symbol may be employed when
the writer wishes to explicitly mark the phonetic glottal stop or to
disambiguate syllable boundaries in connected text
(e.g., $\langle\mathrm{saghang}\rangle$ versus
$\langle\mathrm{saang}\rangle$).

\subsection{Vowels and Codas}
\label{sec:vowels-codas}

Thai and Lao have highly similar vowel and coda systems, which allows the two languages to share a largely unified romanization of vowels and codas. Both languages share the same basic set of nine
monophthong qualities with a phonemic length contrast, as well as closely
corresponding diphthong systems and major coda
series~\cite{iwasaki2005reference,enfield2007grammar}.
The main difference is that Thai
additionally shows marginal final {\ipafont /s/} and {\ipafont /f/},
primarily in loanwords, whereas fricative codas are absent from the Lao
inventory~\cite{kenstowicz2006loanword,enfield2007grammar}. The proposed romanizations are summarized below.

\begin{table}[ht]
\centering
\caption{Monophthongs and their romanization}
\label{tab:monophthongs}
\begin{tabular}{cccc}
\toprule
\multicolumn{2}{c}{Short} & \multicolumn{2}{c}{Long} \\
\cmidrule(lr){1-2}\cmidrule(lr){3-4}
IPA & Roman & IPA & Roman \\
\midrule
{\ipafont /i/}
  & $\langle\mathrm{i}\rangle$
  & {\ipafont /iː/}
  & $\langle\mathrm{ii}\rangle$ \\

{\ipafont /ɯ/}
  & $\langle\mathrm{eo}\rangle$
  & {\ipafont /ɯː/}
  & $\langle\mathrm{eoo}\rangle$ \\

{\ipafont /u/}
  & $\langle\mathrm{u}\rangle$
  & {\ipafont /uː/}
  & $\langle\mathrm{uu}\rangle$ \\

{\ipafont /e/}
  & $\langle\mathrm{ea}\rangle$
  & {\ipafont /eː/}
  & $\langle\mathrm{eaa}\rangle$ \\

{\ipafont /ɤ/}
  & $\langle\mathrm{e}\rangle$
  & {\ipafont /ɤː/}
  & $\langle\mathrm{ee}\rangle$ \\

{\ipafont /o/}
  & $\langle\mathrm{oe}\rangle$
  & {\ipafont /oː/}
  & $\langle\mathrm{oee}\rangle$ \\

{\ipafont /ɛ/}
  & $\langle\mathrm{ae}\rangle$
  & {\ipafont /ɛː/}
  & $\langle\mathrm{aee}\rangle$ \\

{\ipafont /ɔ/}
  & $\langle\mathrm{o}\rangle$
  & {\ipafont /ɔː/}
  & $\langle\mathrm{oo}\rangle$ \\

{\ipafont /a/}
  & $\langle\mathrm{a}\rangle$
  & {\ipafont /aː/}
  & $\langle\mathrm{aa}\rangle$ \\
\bottomrule
\end{tabular}
\end{table}

\begin{table}[ht]
\centering
\caption{Diphthongs and their romanization}
\label{tab:diphthongs}
\begin{tabular}{lll}
\toprule
IPA & Roman & Status \\
\midrule
{\ipafont /ia/}
  & $\langle\mathrm{ia}\rangle$
  & more restricted \\

{\ipafont /iːa/}
  & $\langle\mathrm{iia}\rangle$
  & regular \\

{\ipafont /ua/}
  & $\langle\mathrm{ua}\rangle$
  & more restricted \\

{\ipafont /uːa/}
  & $\langle\mathrm{uua}\rangle$
  & regular \\

{\ipafont /ɯa/}
  & $\langle\mathrm{eoa}\rangle$
  & more restricted \\

{\ipafont /ɯːa/}
  & $\langle\mathrm{eooa}\rangle$
  & regular \\
\bottomrule
\end{tabular}
\end{table}

\begin{table}[ht]
\centering
\caption{Coda consonants and their romanization}
\label{tab:codas}
\begin{tabular}{@{}ccc@{}}
\toprule
\textbf{IPA} & \textbf{Thai Rom.} & \textbf{Lao Rom.} \\
\midrule
{\ipafont /j/} & \multicolumn{2}{c}{$\langle\mathrm{i}\rangle$} \\
{\ipafont /w/} & \multicolumn{2}{c}{$\langle\mathrm{u}\rangle$} \\
{\ipafont /p/} & \multicolumn{2}{c}{$\langle\mathrm{p}\rangle$} \\
{\ipafont /t/} & \multicolumn{2}{c}{$\langle\mathrm{t}\rangle$} \\
{\ipafont /k/} & \multicolumn{2}{c}{$\langle\mathrm{k}\rangle$} \\
{\ipafont /m/} & \multicolumn{2}{c}{$\langle\mathrm{m}\rangle$} \\
{\ipafont /n/} & \multicolumn{2}{c}{$\langle\mathrm{n}\rangle$} \\
{\ipafont /ŋ/} & \multicolumn{2}{c}{$\langle\mathrm{ng}\rangle$} \\
{\ipafont /f/} & $\langle\mathrm{f}\rangle$ & --- \\
{\ipafont /s/} & $\langle\mathrm{s}\rangle$ & --- \\
\bottomrule
\end{tabular}
\end{table}

\noindent\textbf{Coda romanization.}
Coda romanization follows Jyutping as its primary model. Cantonese possesses a coda
inventory particularly close to that required here, and Jyutping writes
its eight major stop, nasal, and glide codas as
$\langle\mathrm{p}\rangle$, $\langle\mathrm{t}\rangle$,
$\langle\mathrm{k}\rangle$, $\langle\mathrm{m}\rangle$,
$\langle\mathrm{n}\rangle$, $\langle\mathrm{ng}\rangle$,
$\langle\mathrm{i}\rangle$, and $\langle\mathrm{u}\rangle$,
corresponding respectively to {\ipafont /p, t, k, m, n, ŋ, j, w/}
in phonological terms~\cite{bauer1997cantonese,matthews2011cantonese}.
The proposed Thai--Lao scheme therefore adopts these coda
spellings directly, following a phonetic correspondence with Jyutping for the shared coda categories.

Pinyin independently supports part of this mapping, using $\langle\mathrm{n}\rangle$ and $\langle\mathrm{ng}\rangle$ for nasal codas and $\langle\mathrm{i}\rangle$ and $\langle\mathrm{u}\rangle$ in finals containing palatal and labial offglides~\cite{norman1988chinese}.

\noindent\textbf{Vowel length marking.}
Phonemic vowel length is represented by graphemic doubling, a widely
attested strategy in Latin-script orthographies. For monographic vowel
symbols, the vowel letter is doubled directly, as in
$\langle\mathrm{i}\rangle$--$\langle\mathrm{ii}\rangle$ and
$\langle\mathrm{a}\rangle$--$\langle\mathrm{aa}\rangle$.
For digraphic short-vowel symbols, only the second letter is doubled,
yielding, for example, $\langle\mathrm{eo}\rangle$--$\langle\mathrm{eoo}\rangle$,
$\langle\mathrm{ae}\rangle$--$\langle\mathrm{aee}\rangle$, and
$\langle\mathrm{oe}\rangle$--$\langle\mathrm{oee}\rangle$.

\noindent\textbf{\boldmath$\langle\mathrm{e}\rangle$ for
{\ipafont /ɤ/}, and $\langle\mathrm{ae}\rangle$ and
$\langle\mathrm{ea}\rangle$ for the front mid vowels.}
The representation of {\ipafont /ɤ/} as
$\langle\mathrm{e}\rangle$ follows Pinyin, where
$\langle\mathrm{e}\rangle$ is commonly associated with the back
unrounded vowel {\ipafont [ɤ]}
~\cite{norman1988chinese,Duanmu2007}.
Jyutping instead associates $\langle\mathrm{e}\rangle$ primarily with
front mid vowel qualities such as {\ipafont [ɛː]} and
{\ipafont [e]}~\cite{bauer1997cantonese,matthews2011cantonese}.
Where the two conventions conflict, the present scheme gives greater
weight to Pinyin because of its broader user base and international
standardization~\cite{iso7098,norman1988chinese}.

With $\langle\mathrm{e}\rangle$ reserved for {\ipafont /ɤ/},
distinct symbols are required for the front mid vowels
{\ipafont /ɛ/} and {\ipafont /e/}. The digraphs
$\langle\mathrm{ae}\rangle$ and $\langle\mathrm{ea}\rangle$
are both familiar from Latin-script spellings of front or mid vowels:
the Revised Romanization of Korean uses
$\langle\mathrm{ae}\rangle$ for \ko{ㅐ}
~\cite{nikl2000romanization}, while
$\langle\mathrm{ea}\rangle$ occurs in English front-vowel spellings
such as \textit{head}~\cite{carney1994survey}.

The internal ordering of the two digraphs provides an additional
mnemonic cue. Since {\ipafont /ɛ/} is the more open of the two vowels,
its symbol begins with $\langle\mathrm{a}\rangle$, whereas the less open
{\ipafont /e/} begins with $\langle\mathrm{e}\rangle$.
This ordering accords with the familiar association of
$\langle\mathrm{a}\rangle$ with a more open vowel quality than
$\langle\mathrm{e}\rangle$, making the contrast between
$\langle\mathrm{ae}\rangle$ and $\langle\mathrm{ea}\rangle$
visually motivated rather than arbitrary.

\noindent\textbf{\boldmath$\langle\mathrm{eo}\rangle$ for
{\ipafont /ɯ/}.}
Neither Pinyin nor Jyutping provides a direct precedent for
{\ipafont /ɯ/}~\cite{Duanmu2007,bauer1997cantonese,matthews2011cantonese}.
A useful regional precedent is the Revised Romanization of Korean,
in which the Korean high unrounded vowel \ko{ㅡ}, conventionally analyzed as
{\ipafont /ɯ/}, is romanized as
$\langle\mathrm{eu}\rangle$~\cite{sohn1999korean,nikl2000romanization}.
Other nearby precedents are less compatible with the present scheme:
Vietnamese uses the non-ASCII letter \textit{ư}, while Japanese
romanization schemes generally use $\langle\mathrm{u}\rangle$, which is already
assigned to {\ipafont /u/} in the present scheme
~\cite{thompson1965vietnamese,labrune2012japanese}.

Directly adopting Korean $\langle\mathrm{eu}\rangle$, however, would
create a segmentation ambiguity because final
$\langle\mathrm{u}\rangle$ independently represents the glide coda
{\ipafont /w/}. The sequence $\langle\mathrm{eu}\rangle$ could therefore
be parsed either as a single vowel symbol or as
$\langle\mathrm{e}\rangle+\langle\mathrm{u}\rangle$.
To avoid this ambiguity, the Korean-inspired form is modified to
$\langle\mathrm{eo}\rangle$.

The substitution of $\langle\mathrm{o}\rangle$ for
$\langle\mathrm{u}\rangle$ also retains a secondary phonetic and graphic
motivation, since the two letters are conventionally associated with
neighboring back rounded vowel qualities that differ primarily in height
~\cite{ladefoged1996sounds}.
Thus, $\langle\mathrm{eo}\rangle$ preserves the two-letter structure of
the Korean precedent while remaining distinct from a vowel followed by
the productive $\langle\mathrm{u}\rangle$ coda.
Although $\langle\mathrm{eo}\rangle$ has a different sound value in
Jyutping, the present assignment prioritizes the Korean precedent and
unambiguous syllable parsing.

\noindent\textbf{\boldmath$\langle\mathrm{o}\rangle$ for
{\ipafont /ɔ/} and $\langle\mathrm{oe}\rangle$ for
{\ipafont /o/}.}
The representation of {\ipafont /ɔ/} as
$\langle\mathrm{o}\rangle$ follows Jyutping, where
$\langle\mathrm{o}\rangle$ is predominantly associated with the
open-mid back rounded vowel quality {\ipafont [ɔ]}
~\cite{bauer1997cantonese,matthews2011cantonese}.
The present scheme adopts this phonetic correspondence.

Thai and Lao, however, phonemically distinguish
{\ipafont /ɔ/} from {\ipafont /o/}, so a separate symbol is required
for the latter~\cite{iwasaki2005reference,enfield2007grammar}.
The proposed scheme assigns {\ipafont /o/} the digraph
$\langle\mathrm{oe}\rangle$.
Its internal structure is deliberately iconic: the initial
$\langle\mathrm{o}\rangle$ preserves the association with a back
rounded vowel, while the following $\langle\mathrm{e}\rangle$
functions as a cue to the smaller degree of vowel opening.
This motivation is reflected both in the IPA vowel space and in the
internal structure of the present scheme. In the IPA vowel space,
{\ipafont /e/} and {\ipafont /o/} occupy corresponding close-mid
positions, whereas {\ipafont /ɔ/} is open-mid; similarly, in the
present scheme, $\langle\mathrm{e}\rangle$ represents
{\ipafont /ɤ/}, which is likewise close-mid
~\cite{ladefoged1996sounds}.
Thus, relative to $\langle\mathrm{o}\rangle$ for {\ipafont /ɔ/},
the added $\langle\mathrm{e}\rangle$ serves as a cue to reduced vowel
openness, while the initial $\langle\mathrm{o}\rangle$ retains the
backness and rounding cues, yielding $\langle\mathrm{oe}\rangle$
for {\ipafont /o/}.

\noindent\textbf{Compositional romanization of diphthongs.}
The three diphthongal patterns are romanized compositionally using
the symbols already assigned to their component vowels.
No additional diphthong-specific symbols are introduced.
The initial element retains the vowel-quality and length representation of
the monophthong system, while the final {\ipafont /a/} is represented
by $\langle\mathrm{a}\rangle$. Thus, for example,
{\ipafont /iːa/} is written $\langle\mathrm{iia}\rangle$, combining
$\langle\mathrm{ii}\rangle$ with final $\langle\mathrm{a}\rangle$.
The remaining diphthongs follow the same compositional rule.

The long forms constitute the regular diphthongal series in Thai,
whereas the corresponding short forms are highly restricted:
short {\ipafont /ia/} and {\ipafont /ua/} occur only in marginal
lexical items, while {\ipafont /ɯa/} is reported as unattested in
some descriptions~\cite{diller2008thai}.
Lao, by contrast, distinguishes short and long forms for all three
diphthongal patterns, although the short series is lexically much more
restricted than the regular long forms
~\cite{enfield2007grammar,lew2014lao}.
The proposed scheme therefore retains all six combinations for
systematic completeness, with the short forms available wherever the
corresponding short diphthongal categories are required.

\subsection{Tones}
\label{sec:tones}

Tone requires a somewhat different treatment from segmental phonemes.
Whereas onset, nuclear vowel, and coda components can generally be
represented directly by Latin letters or letter sequences, tone is
suprasegmental and therefore requires an additional notational mechanism
~\cite{yip2002tone}.

The tone design of the present Thai--Lao scheme builds on the Sinitic Romanization Ecosystem proposed by Zhang et al.~\cite{zhang2026iscslp}, particularly its dual-track tone design for Mandarin, Cantonese, and more broadly Sinitic romanization. That work introduced a
dual-track tone design combining a \textit{tone-value} representation
for phonetic correspondence with a \textit{tone-category}
representation for historical-phonological correspondence. The
motivation for separating the two is that tone values may change
substantially over time, making synchronic pitch realization
non-transparent with respect to historical tone category
~\cite{zhang2026iscslp,endo2016}. The dual-track design therefore allows
synchronic phonetic and historical-phonological information to be
represented independently rather than forcing both into a single tone
notation.

The same work also showed that the linguistically informative
multi-character dual-track tone design is not necessarily optimal as a
speech-recognition target compared with a compact single-digit tone
romanization design~\cite{zhang2026iscslp}. The present Thai--Lao scheme
therefore retains both ideas from the earlier design: a compact
single-digit tone notation and the optional dual-track representations.
The compact notation is adopted as the default. Each synchronic tone is
assigned a single language-specific digit from
$\langle\mathrm{1}\rangle$ to $\langle\mathrm{5}\rangle$, producing a
short representation that is visually and orthographically similar to
the familiar high-readability single-digit tone conventions of Pinyin
and Jyutping and is also well suited to computational use.

The dual-track design is retained as an optional analytic notation for
contexts in which phonetic or historical-phonological information needs
to be made explicit. Its tone-value track uses Chao's five-level
notation to represent synchronic pitch transparently and therefore
directly serves phonetic correspondence, while its tone-category track
uses the Hierarchical Tone-Category Notation (HTCN) to represent
historical-phonological correspondence. Both tracks generally require
multiple characters for a single tone and are therefore less consistent
with the compact single-digit tone-marking style familiar from Pinyin
and Jyutping. Accordingly, they are not used as the default tone
notation, but are available when explicit representation of phonetic
realization or historical-phonological correspondence is required. Because the default compact notation is derived partly from the
tone-category track, we first introduce the two optional tracks and
then turn to the compact notation.

\noindent\textbf{Optional dual-track tone notation.} Within this optional dual-track notation, Chao's five-level system
represents relative pitch height on a scale from 1 (lowest) to 5
(highest)~\cite{chao1930tone,rattanasone2013compare}. Sequences of
these digits are used directly as the tone-value representation of
the synchronic pitch contours of Thai and Lao tones.

The tone-category track requires a language-specific adaptation of
HTCN to the conventional historical analysis of Tai tones. In its
original formulation for Sinitic languages, HTCN encodes historical
tone categories hierarchically~\cite{zhang2026iscslp}. The first level
uses $\langle\mathrm{1}\rangle$--$\langle\mathrm{4}\rangle$ for the
traditional Middle Chinese level (\textit{píng} 平), rising
(\textit{shǎng} 上), departing (\textit{qù} 去), and entering
(\textit{rù} 入) categories, respectively. A second-level
$\langle\mathrm{A}\rangle$--$\langle\mathrm{B}\rangle$ distinction
represents the yīn--yáng register split, while a third-level
$\langle\mathrm{a}\rangle$--$\langle\mathrm{b}\rangle$ distinction
may be added where a finer historical subdivision is required, as in
the upper and lower yīn entering tones of Yue Chinese
~\cite{yue2015yue}. Thus, yīn level and yáng level are represented as
$\langle\mathrm{1A}\rangle$ and $\langle\mathrm{1B}\rangle$, while
the upper and lower yīn entering categories are represented as
$\langle\mathrm{4Aa}\rangle$ and $\langle\mathrm{4Ab}\rangle$.

Thai and Lao, however, belong to the Tai branch of the Kra--Dai
family rather than to the Sinitic family, and their historical tone
categories are therefore not descendants of the Middle Chinese
level, rising, departing, and entering categories. Comparative Tai
reconstruction conventionally distinguishes the historical categories
{\ipafont *A}, {\ipafont *B}, {\ipafont *C}, and
{\ipafont *D}~\cite{gedney1972checklist,pittayaporn2009prototai}.
The first three are associated with unchecked syllables, whereas
{\ipafont *D} is the conventional category for checked syllables
ending in obstruents and is therefore comparable in distribution to
the entering-tone category of traditional Chinese phonology
~\cite{pittayaporn2009prototai}. Within {\ipafont *D}, vowel length
provides a further historical distinction:
{\ipafont *DS} denotes syllables with short vowels, whereas
{\ipafont *DL} denotes those with long vowels
~\cite{gedney1972checklist,pittayaporn2009prototai}.

Gedney's tone-box framework organizes Tai historical tonology into
five columns, {\ipafont *A}, {\ipafont *B}, {\ipafont *C},
{\ipafont *DS}, and {\ipafont *DL}, cross-classified by four
historical onset classes conventionally represented as rows 1--4
~\cite{gedney1972checklist}. These correspond respectively to
historically voiceless friction initials, voiceless unaspirated
initials, glottal or glottalized initials, and voiced initials.
The resulting twenty cells are designated
{\ipafont A1--A4}, {\ipafont B1--B4},
{\ipafont C1--C4}, {\ipafont DS1--DS4}, and
{\ipafont DL1--DL4}. Individual Tai languages generally merge
multiple cells into a smaller number of synchronic lexical tones,
while the tone-box notation preserves their distinct historical
sources.

Table~\ref{tab:tonebox-reflexes} summarizes the relevant Gedney-box
sources of the synchronic Thai and Lao tones. The Thai correspondences
follow the conventional Gedney-box analysis, while the Lao
correspondences follow the modern five-tone analysis reported by
Osatananda and subsequently presented in tone-box form by
Pornpottanamas~\cite{gedney1972checklist,liao2023tonal,
pornpottanamas2023lao}.

\begin{table}[ht]
\centering
\caption{Historical tone-box sources of Thai and Lao tones}
\label{tab:tonebox-reflexes}
\footnotesize
\begin{tabular}{@{}lll@{}}
\toprule
\textbf{Historical cell} & \textbf{Thai} & \textbf{Lao} \\
\midrule
A1     & rising 24       & low-rising 24 \\
A2--3  & mid 33          & low-rising 24 \\
A4     & mid 33          & mid-rising 35 \\
B1--3  & low 21          & mid-level 33 \\
B4     & falling 41      & mid-level 33 \\
C1     & falling 41      & mid-falling 31 \\
C2--3  & falling 41      & high-falling 42 \\
C4     & high 45         & high-falling 42 \\
DL1--3 & low 21          & mid-falling 31 \\
DL4    & falling 41      & high-falling 42 \\
DS1--3 & low 21          & mid-rising 35 \\
DS4    & high 45         & mid-level 33 \\
\bottomrule
\end{tabular}
\end{table}

The Gedney tone box and the tone-category track of HTCN are
structurally compatible in that both encode progressively finer
historical distinctions~\cite{zhang2026iscslp,gedney1972checklist}.
Although Gedney's system is conventionally organized as a tone box
rather than a linear hierarchy, its historical dimensions can be
expressed naturally within the hierarchical format of HTCN.

More importantly, the correspondence between the Sinitic and Tai
tone-category systems has long been recognized in comparative
scholarship. Sinitic level (\textit{píng} 平), rising
(\textit{shǎng} 上), departing (\textit{qù} 去), and entering
(\textit{rù} 入) correspond respectively to Proto-Tai
{\ipafont *A, *C, *B}, and {\ipafont *D}
~\cite{li1977handbook,sagart1988glottalised,bauer1996tai,
luo2008sinotai,liao2023tonal}. The historical interpretation of this
relationship varies across studies, including accounts involving
lexical correspondence, language contact and areal diffusion, and
comparative tonal development. On the basis of this established
tone-category correspondence, and in keeping with the present
scheme's aim of increasing correspondence and compatibility with Sinitic romanization for users familiar with Pinyin or Jyutping, we map the Proto-Tai categories and
Gedney tone-box notation onto HTCN. This adaptation is intended to maintain interoperability with the previously proposed Sinitic Romanization Ecosystem~\cite{zhang2026iscslp}, rather than to impose a Sinitic analysis on Tai tonology; the underlying Tai categories remain those of the conventional Gedney framework.

\begin{table}[ht]
\centering
\caption{Tai--HTCN tone-category correspondence}
\label{tab:tai-htcn-mapping}
\footnotesize
\setlength{\tabcolsep}{3pt}
\begin{tabular}{@{}lcl@{}}
\toprule
\textbf{\shortstack{Comparative Tai\\notation}} &
\textbf{\shortstack{Adapted\\HTCN}} &
\textbf{\shortstack{Sinitic tone\\category}} \\
\midrule

\multicolumn{3}{@{}l}{\textit{Proto-Tai tone category}} \\
{\ipafont *A} & $\langle\mathrm{1}\rangle$ &
level (\textit{píng} 平) \\
{\ipafont *C} & $\langle\mathrm{2}\rangle$ &
rising (\textit{shǎng} 上) \\
{\ipafont *B} & $\langle\mathrm{3}\rangle$ &
departing (\textit{qù} 去) \\
{\ipafont *D} & $\langle\mathrm{4}\rangle$ &
entering (\textit{rù} 入) \\

\addlinespace
\multicolumn{3}{@{}l}{\textit{Gedney tone-box onset class}} \\
Row 2 & $\langle\mathrm{A}\rangle$ & --- \\
Row 4 & $\langle\mathrm{B}\rangle$ & --- \\
Row 1 & $\langle\mathrm{C}\rangle$ & --- \\
Row 3 & $\langle\mathrm{D}\rangle$ & --- \\

\addlinespace
\multicolumn{3}{@{}l}{\textit{Checked-syllable vowel length}} \\
S in {\ipafont *DS} & $\langle\mathrm{a}\rangle$ & --- \\
L in {\ipafont *DL} & $\langle\mathrm{b}\rangle$ & --- \\

\bottomrule
\end{tabular}
\end{table}

Table~\ref{tab:tai-htcn-mapping} summarizes the proposed mapping
from conventional comparative Tai notation to the adapted HTCN. The
mapping operates at three hierarchical levels. Proto-Tai tone
categories determine the initial HTCN digit; the four Gedney tone-box
onset classes determine the following uppercase letter; and, within
the checked {\ipafont *D} category, the short--long distinction
between {\ipafont *DS} and {\ipafont *DL} determines the final
lowercase letter. The resulting notation therefore preserves the
historical information of the conventional Tai tone box while
expressing it within the same hierarchical format as the Sinitic HTCN.

\noindent\textbf{\boldmath Second HTCN level:
$\langle\mathrm{A}\rangle$--$\langle\mathrm{D}\rangle$ for Gedney classes.}
Gedney row 2 is characterized by historically voiceless unaspirated
initials such as {\ipafont */p/, */t/, */k/}
~\cite{gedney1972checklist}. In the Sinitic HTCN,
$\langle\mathrm{A}\rangle$ represents the yīn register split, which
developed primarily from historically voiceless initials in Middle
Chinese~\cite{yip2002tone,bao1999tone}. Among the Gedney classes,
row 2 provides the most direct correspondence with this core
voiceless series and is therefore mapped to
$\langle\mathrm{A}\rangle$.

Gedney row 4, characterized by historically voiced initials,
corresponds directly to the historical voicing condition of the
Sinitic yáng register and is therefore mapped to
$\langle\mathrm{B}\rangle$
~\cite{gedney1972checklist,yip2002tone,bao1999tone}.

Gedney row 1 comprises historically voiceless friction initials,
including voiceless or preaspirated sonorants such as
{\ipafont */hm/, */hn/, */hɲ/, */hŋ/, */hl/, */hw/, */hr/, */hj/},
as well as other voiceless friction and aspirated initials
~\cite{gedney1972checklist,liao2023tonal}.
This broader class has no direct counterpart in the Middle Chinese
initial classes underlying the Sinitic yīn--yáng distinction.
After $\langle\mathrm{A}\rangle$ and $\langle\mathrm{B}\rangle$
are assigned to the two most directly comparable classes, we assign
the next uppercase symbol, $\langle\mathrm{C}\rangle$, to Gedney
row 1.

Gedney row 3 comprises glottal or glottalized initials, including
{\ipafont */ʔ/} and preglottalized stops
~\cite{gedney1972checklist}. As this class likewise has no direct
counterpart in the Sinitic HTCN register distinction, it is assigned
the next available uppercase symbol, $\langle\mathrm{D}\rangle$,
completing the Tai-specific second-level series
$\langle\mathrm{A}\rangle$--$\langle\mathrm{D}\rangle$.

\noindent\textbf{\boldmath$\langle\mathrm{a}\rangle$ and
$\langle\mathrm{b}\rangle$ for {\ipafont S} and {\ipafont L}.}
Thai and Lao, as closely related Tai languages within the Kra--Dai
family, share the comparative-Tai distinction between
{\ipafont *DS} and {\ipafont *DL}, in which the checked
{\ipafont *D} category is subdivided according to short and long
vowel length, respectively~\cite{gedney1972checklist,li1977handbook}.
A closely parallel subdivision is found in Cantonese. Its yīn entering
tone is conventionally divided into upper yīn entering and lower yīn
entering, with the former associated primarily with short vowels and
the latter with long vowels~\cite{bauer1996tai,bauer2003cantonese}.
Previous studies have explicitly related this length-conditioned
Cantonese checked-tone split to Tai or more broadly Kra--Dai
influence, variously interpreting the relationship in terms of a Tai
substratum, historical language contact, or areal diffusion
~\cite{bauer1996tai,bauer2003cantonese,desousa2015southern,
liao2023tonal}.

In the original Sinitic HTCN, the third-level symbols
$\langle\mathrm{a}\rangle$ and $\langle\mathrm{b}\rangle$ represent
the upper and lower subdivisions of yīn entering tone, respectively
~\cite{zhang2026iscslp}. Since the upper category is associated with
short vowels and the lower category with long vowels, the adapted Tai
HTCN extends the same distinction to the corresponding vowel-length
subdivision of {\ipafont *D}: $\langle\mathrm{a}\rangle$ represents
S in {\ipafont *DS}, while $\langle\mathrm{b}\rangle$ represents L
in {\ipafont *DL}. This mapping therefore preserves not only the
hierarchical structure of HTCN but also the close historical and
areal parallel between the checked-tone systems of Tai and Cantonese.

The complete conversion of the twenty Gedney tone-box cells under
these mapping rules is shown in Table~\ref{tab:gedney-htcn}. Each
HTCN form combines the first-level tone-category digit with the
second-level Gedney-class symbol and, for checked {\ipafont *D}
categories, the third-level vowel-length symbol.

\begin{table}[ht]
\centering
\caption{HTCN representations of Gedney tone-box cells}
\label{tab:gedney-htcn}
\footnotesize
\setlength{\tabcolsep}{4pt}
\begin{tabular}{@{}cccccc@{}}
\toprule
\textbf{Row} &
\textbf{{\ipafont *A}} &
\textbf{{\ipafont *B}} &
\textbf{{\ipafont *C}} &
\textbf{{\ipafont *DS}} &
\textbf{{\ipafont *DL}} \\
\midrule
1 &
$\langle\mathrm{1C}\rangle$ &
$\langle\mathrm{3C}\rangle$ &
$\langle\mathrm{2C}\rangle$ &
$\langle\mathrm{4Ca}\rangle$ &
$\langle\mathrm{4Cb}\rangle$ \\

2 &
$\langle\mathrm{1A}\rangle$ &
$\langle\mathrm{3A}\rangle$ &
$\langle\mathrm{2A}\rangle$ &
$\langle\mathrm{4Aa}\rangle$ &
$\langle\mathrm{4Ab}\rangle$ \\

3 &
$\langle\mathrm{1D}\rangle$ &
$\langle\mathrm{3D}\rangle$ &
$\langle\mathrm{2D}\rangle$ &
$\langle\mathrm{4Da}\rangle$ &
$\langle\mathrm{4Db}\rangle$ \\

4 &
$\langle\mathrm{1B}\rangle$ &
$\langle\mathrm{3B}\rangle$ &
$\langle\mathrm{2B}\rangle$ &
$\langle\mathrm{4Ba}\rangle$ &
$\langle\mathrm{4Bb}\rangle$ \\
\bottomrule
\end{tabular}
\end{table}

When a synchronic Thai or Lao tone has more than one historical
Gedney tone-box source, the HTCN representations of the individual
source cells are concatenated linearly in their conventional tone-box
order. For example, the Thai mid tone 33 derives from Gedney cells
A2, A3, and A4, whose HTCN representations are
$\langle\mathrm{1A}\rangle$,
$\langle\mathrm{1D}\rangle$, and
$\langle\mathrm{1B}\rangle$, respectively. The tone is therefore
represented as $\langle\mathrm{1A1D1B}\rangle$.

The resulting tone-category representations and corresponding tone-value
representations for Thai and Lao are summarized in
Table~\ref{tab:thai-lao-tones}. The table also includes the default
compact single-digit tone notation, whose assignment is introduced
below.

\begin{table}[ht]
\centering
\caption{Compact, tone-value, and HTCN representations of Thai and Lao tones}
\label{tab:thai-lao-tones}
\footnotesize
\setlength{\tabcolsep}{3pt}
\begin{tabular}{@{}lccc@{}}
\toprule
\textbf{Tone} & \textbf{Compact} & \textbf{Value} & \textbf{HTCN} \\
\midrule

\multicolumn{4}{@{}l}{\textit{Thai}} \\
Mid     & 1 & 33 & $\langle\mathrm{1A1D1B}\rangle$ \\
Rising  & 2 & 24 & $\langle\mathrm{1C}\rangle$ \\
High    & 3 & 45 & $\langle\mathrm{2B4Ba}\rangle$ \\
Falling & 4 & 41 & $\langle\mathrm{3B2C4Bb}\rangle$ \\
Low     & 5 & 21 & $\langle\mathrm{3C3A3D4Ca4Aa4Da4Cb4Ab4Db}\rangle$ \\

\addlinespace
\multicolumn{4}{@{}l}{\textit{Lao}} \\
Low-rising   & 1 & 24 & $\langle\mathrm{1C1A1D}\rangle$ \\
Mid-rising   & 2 & 35 & $\langle\mathrm{1B4Ca4Aa4Da}\rangle$ \\
High-falling & 3 & 42 & $\langle\mathrm{2A2D2B4Bb}\rangle$ \\
Mid-falling  & 4 & 31 & $\langle\mathrm{2C4Cb4Ab4Db}\rangle$ \\
Mid-level    & 5 & 33 & $\langle\mathrm{3C3A3D3B4Ba}\rangle$ \\

\bottomrule
\end{tabular}
\end{table}

\noindent\textbf{Compact single-digit tone notation.}
The default tone numbering draws on the conventional single-digit
ordering of tone categories in Chinese dialectology, in which yīn level,
yáng level, yīn rising, yáng rising, yīn departing, yáng departing,
yīn entering, and yáng entering are ordered sequentially. In HTCN terms,
these correspond to
$\langle\mathrm{1A}\rangle$,
$\langle\mathrm{1B}\rangle$,
$\langle\mathrm{2A}\rangle$,
$\langle\mathrm{2B}\rangle$,
$\langle\mathrm{3A}\rangle$,
$\langle\mathrm{3B}\rangle$,
$\langle\mathrm{4A}\rangle$, and
$\langle\mathrm{4B}\rangle$, respectively.
For the adapted Tai HTCN, this ordering is extended by ordering the
second-level symbols as
$\langle\mathrm{A}\rangle<
\langle\mathrm{B}\rangle<
\langle\mathrm{C}\rangle<
\langle\mathrm{D}\rangle$
within each first-level category, with
$\langle\mathrm{a}\rangle<
\langle\mathrm{b}\rangle$
at the third level where applicable.

Because most synchronic Thai and Lao tones merge multiple historical
tone-box sources, they cannot generally be identified with a single
HTCN category. We therefore use the earliest constituent HTCN category
under the ordering above as the ordering key for each synchronic tone.
The five tones of each language are then ranked by these keys and
assigned the compact digits
$\langle\mathrm{1}\rangle$--$\langle\mathrm{5}\rangle$
in that order. This procedure preserves the conventional Sinitic-style
tone-category ordering as far as possible while assigning exactly one
compact digit to each synchronic tone.

For Thai, the ordering keys are
$\langle\mathrm{1A}\rangle$ for the mid tone,
$\langle\mathrm{1C}\rangle$ for the rising tone,
$\langle\mathrm{2B}\rangle$ for the high tone,
$\langle\mathrm{2C}\rangle$ for the falling tone, and
$\langle\mathrm{3A}\rangle$ for the low tone, yielding compact tone
numbers 1--5 in that order. For Lao, the corresponding ordering keys are
$\langle\mathrm{1A}\rangle$,
$\langle\mathrm{1B}\rangle$,
$\langle\mathrm{2A}\rangle$,
$\langle\mathrm{2C}\rangle$, and
$\langle\mathrm{3A}\rangle$, yielding 1 for low-rising, 2 for
mid-rising, 3 for high-falling, 4 for mid-falling, and 5 for mid-level.

\section{Conclusion}

This paper proposes a phonemically comprehensive, ASCII-only romanization scheme that treats Thai and Lao within a unified cross-lingual framework. The scheme represents segmental contrasts, vowel length, and lexical tone
while maintaining a highly readable alphanumeric inventory and
one-symbol-one-phoneme transparency. Shared or closely corresponding
Thai and Lao phonemes are represented systematically, while Pinyin and Jyutping conventions are incorporated where they provide
appropriate phonetic correspondences for users familiar with Pinyin or Jyutping.
Historical-phonological correspondence is preserved where compatible
with synchronic phonetic transparency. For tone, the scheme combines a
compact single-digit default notation with optional tone-value and
tone-category representations, allowing practical use and more explicit
phonetic or historical-phonological representations within the same
framework.

This work focuses on scheme design rather than empirical evaluation. Its proposed benefits for readability, learnability, and computational processing remain to be empirically tested. Future work should evaluate the scheme across user populations, particularly those familiar with Pinyin or Jyutping, and in downstream speech and language processing tasks, including cross-lingual Thai--Lao transfer.

\section{Generative AI Use Disclosure} 
Generative AI tools were used for language editing and polishing of the manuscript. All scientific content, analyses, design decisions, and conclusions were reviewed and approved by the authors, who take full responsibility for the final manuscript.

\bibliographystyle{IEEEtran}
\bibliography{mybib}

\end{document}